\documentclass[pdflatex,iicol, sn-mathphys]{sn-jnl}

\usepackage[utf8]{inputenc}
\usepackage[T1]{fontenc}
\usepackage{csquotes}
\usepackage{multirow}

\jyear{2023}%

\theoremstyle{thmstyleone}%
\theoremstyle{thmstyletwo}%

\theoremstyle{thmstylethree}%

\begin{document}

\title[HSR and the LION Dataset]{Handwritten Stenography Recognition and the LION Dataset}


\author*[1]{\fnm{Raphaela} \sur{Heil}}\email{raphaela.heil@it.uu.se}
\author*[2,3]{\fnm{Malin} \sur{Nauwerck}\email{malin.nauwerck@barnboksinstitutet.se}}

\affil[1]{\orgdiv{Department of Information Technology}, \orgname{Uppsala University}, \orgaddress{\street{Lägerhyddsvägen 1}, \city{Uppsala}, \postcode{75237}, \country{Sweden}}, ORCID: 0000-0002-5010-9149}
\affil[2]{\orgname{The Swedish Institute for Children's Books}, \orgaddress{\street{Odengatan 61}, \city{Stockholm}, \postcode{11322}, \country{Sweden}}}
\affil[3]{\orgdiv{Department of Literature and Rhetoric}, \orgname{Uppsala University}, \orgaddress{\street{Thunbergsvägen 3P}, \city{Uppsala}, \postcode{75238}, \country{Sweden}}, ORCID: 0000-0002-4834-3761}







\abstract{\textbf{Purpose:} In this paper, we establish a baseline for handwritten stenography recognition, using the novel LION dataset, and investigate the impact of including selected aspects of stenographic theory into the recognition process. We make the LION dataset publicly available with the aim of encouraging future research in handwritten stenography recognition. 

\textbf{Methods:} A state-of-the-art text recognition model is trained to establish a baseline. Stenographic domain knowledge is integrated by applying four different encoding methods that transform the target sequence into representations, which approximate selected aspects of the writing system. Results are further improved by integrating a pre-training scheme, based on synthetic data. 

\textbf{Results:} The baseline model achieves an average test character error rate (CER) of 29.81\% and a word error rate (WER) of 55.14\%. Test error rates are reduced significantly by combining stenography-specific target sequence encodings with pre-training and fine-tuning, yielding CERs in the range of 24.5\% - 26\% and WERs of 44.8\% - 48.2\%.

\textbf{Conclusion:} The obtained results demonstrate the challenging nature of stenography recognition. Integrating stenography-specific knowledge, in conjunction with pre-training and fine-tuning on synthetic data, yields considerable improvements. Together with our precursor study on the subject, this is the first work to apply modern handwritten text recognition to stenography. The dataset and our code are publicly available via Zenodo.
}



\maketitle

\section{Introduction}\label{sec:introduction}

\begin{figure*}[ht]
    \centering
    \includegraphics[width=0.8\linewidth]{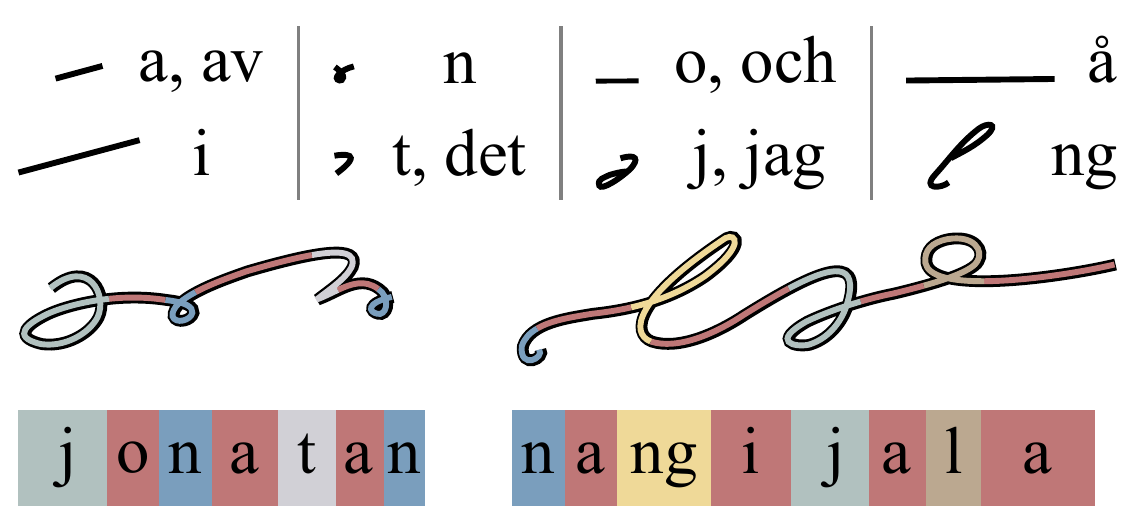}
    \caption{Excerpt from Melin's system. Top: selected characters, shortforms (av, det, och, jag), and an n-gram (ng); bottom: two examples, \textit{jonatan} and \textit{nangijala}, in handwritten stenography; colours visualise the extent of the respective character, as indicated in the transliteration below.}
    \label{fig:enter-label}
\end{figure*}

Stenography, or shorthand, is a method commonly used for speed writing, which manifests itself in many different languages and systems. Because it can be written rapidly, shorthand has traditionally been used by secretaries and reporters, for example in parliament and law courts, for recording testimonies and interviews, or for dictation in business correspondence. A stenographer typically develops a personal style that goes beyond the commonly observed handwriting variations, such as size and slant, and entails an individual bank of abbreviations and innovations. These allow the notetaker some privacy from the uninitiated. 

For Swedish author Astrid Lindgren (1907 -- 2002) who wrote and edited all her literary fiction in the Swedish stenographic system of \textit{Melin}, shorthand provided access to a private intellectual and creative space -- Lindgren’s own version of a “room of one’s own” \cite{nauwerck2022}. Since Lindgren was the editor and publisher of her own books \cite{Bohlund2018}, and did all her editing in her shorthand notepads, the revisions, deletions, and additions they display, constitute the only first hand source to the author's creative process \cite{Andersen2014,tornqvist2015}. Lindgren herself has described her stenographic notes as impossible to interpret. This myth, frequently reproduced in general Lindgren reception, has been dispelled through research within the ongoing digital humanities project "The Astrid Lindgren Code" (2020–2023) \cite{al-code}, where different and mixed methods are applied to approach Lindgren's shorthand \cite{dhnb2022,ibpria}.  

In this work, we study Lindgren's manuscripts from a handwritten text recognition (HTR) perspective. Prior research in stenography recognition has primarily been focused on the English stenographic systems \textit{Gregg's} and \textit{Pitman's} and has, so far, been limited to symbol and individual word recognition \cite{leedham,bayes_pitman,gregg-1916,canny_pitman,philippine_gregg}. Contrary to this, we focus our experiments on line-level transliterations and present a first baseline, employing a state-of-the-art deep learning model \cite{flor_restricted}. We build upon this baseline by extending the training process with domain knowledge, primarily founded in visual aspects of the Melin system, as well as synthetically generated training lines. 

All of our experiments are based on the novel LION dataset, which is published in conjunction with this work. This low-resource corpus consists of a selection of Lindgren's drafts, containing portions of her well-known novel for children, \textit{The Brothers Lionheart}, as well as excerpts from selected other texts. 

Our contributions in this work can be summarised as follows:
\begin{itemize}
    \item the introduction of the novel, stenographic LION dataset (\autoref{sec:dataset})
    \item the establishment of a line-based stenography recognition baseline (\autoref{sec:clean-baseline})
    \item the integration of stenographic domain knowledge into the training process (\autoref{sec:encoding})
    \item further improvement of results via pre-training and fine-tuning (\autoref{sec:pretrain}) 
\end{itemize}

Furthermore, the LION dataset and our code are publicly available via Zenodo (cf. \autoref{sec:data_avail}).

\section{Related Works}
Our work is placed at the intersection of document image processing for stenography and handwritten text recognition. Below we present a summary of relevant, related works for both topics. 

\subsection{Document Image Analysis for Stenography}

Early approaches in document image analysis for stenography date back to the 1980s, for example by Leedham and Downton \cite{leedham}. The research in this field has, so far, been limited to the two main English shorthand systems, Pitman's and Gregg's. Considering the research that has been published within the past ten years, i.e. since 2012, we have identified the following four works. 

In 2012, Htwe et al.\ investigated the use of Bayesian networks as part of the recognition pipeline for Pitman's shorthand \cite{bayes_pitman}.

Zhai et al. proposed a pipeline to perform word-level recognition of Gregg's shorthand \cite{gregg-1916}. For this, they combined a convolutional neural network (CNN), as feature extractor, with a recurrent neural network (RNN) as sequence generator, and refined the generated hypotheses, using a word retrieval module. In addition to their proposed pipeline, Zhai et al. released \textit{Gregg-1916}, a word-level dataset that consists of 15\,711 word images that were extracted from a printed Gregg's shorthand dictionary. 

Montalbo and Barfeh employed Canny edge detection and a CNN to classify 100 commonly-used legal words and phrases, written in Pitman's stenography \cite{canny_pitman}. Following a similar line of research, Padilla et al. investigated the use of Inception-v3 to classify 135 legal terms, written in Gregg's shorthand \cite{philippine_gregg}. 

Lastly, we recently published our precursor study, which uses the LION dataset to investigate the effect of commonly-used HTR data augmentations \cite{ibpria}. Besides our prior work, we are not aware of any literature that investigates the use of state-of-the-art text recognition techniques for any system of stenography. 

\subsection{Handwritten Text Recognition}
Handwritten text recognition approaches (HTR) range from identifying individual symbols (e.g. \cite{symbol-recog}) to page-level recognition (e.g. \cite{origami}). In this work we are focusing on line-based recognition, because our dataset is annotated and segmented at this level. Deep learning-based, line-level text recognition can be further divided into three main categories, based on the employed approaches. 

Firstly, architectures that are trained using the Connectionist Temporal Classification (CTC) loss \cite{ctc_graves,ctc_liwicki}. These approaches generally combine a CNN, with an RNN, such as Gated Recurrent Units (GRU) \cite{gru}. Recent examples for this approach are the works by Neto et al.\ \cite{flor_restricted}, Puigcerver \cite{Puigcerver}, and Bluche and Messina \cite{messina}.

Secondly, sequence-to-sequence (seq2seq) methods that use similar network configurations as the aforementioned approach but add an additional RNN to decode the output sequence. Approaches in this category have for example been proposed by Michael et al.\ \cite{s2s-eval-labahn} and Chowdhury et al.\ \cite{Chowdhury}.

Thirdly, and most recently, transformer-based \cite{transformer} approaches have for example been proposed by Wick et al.\ \cite{rostock-transformer}, Kang et al.\ \cite{kang-transformer}, and Barrere and colleagues \cite{light-transformer}. 

In this work, we focus on CTC-based models, as these have been shown to generally perform well, especially in low-resource settings \cite{flor_restricted,rostock-transformer,kang-transformer}, as is the case for the LION dataset. 

\section{Dataset}\label{sec:dataset}
In this section, we introduce the novel \textit{LION} dataset, which is the first of its kind in several regards. Firstly, it is the first dataset containing a portion of Astrid Lindgren's original drafts and handwriting. Secondly, it is the first to present text, written in the Swedish stenography system Melin. Finally, it is the first publicly available dataset, covering a substantial amount of handwritten lines in any kind of stenographic system.


\subsection{Interdisciplinary Context}

In the following, we contextualise the LION dataset as an object for digital humanities and literary studies, and discuss briefly to which extent the dataset can be considered representative for both Melin shorthand and shorthand in general. We present characteristics of Lindgren's vocabulary and style, and provide selected examples of challenges connected to inconsistencies in Lindgren's shorthand writing.

\subsubsection{Mixed Methods}
Preserved stenographic material can be of varying historical or cultural significance, but notable examples of canonised authors who have made use of shorthand in their writing process include for example Fyodor Dostoyevsky \cite{Andrianova2019} and Charles Dickens \cite{bowles2018,dickens-code}. The providing of access to Astrid Lindgren's shorthand manuscripts is motivated by their status as prominent cultural heritage, and finds further relevance from the perspectives of children's literature, book and media history, and textual and genetic criticism. The latter means entering \enquote{the workshop} \cite{hay2004genetic} of the writer and drawing attention to the labour, craftsmanship, and dynamics of the creative process. Usually it involves the process of organising and making accessible the documents that precede a book's publication, which is achieved by compiling and deciphering relevant documents, establishing a chronological order, and then transcribing and editing the texts \cite{van-hulle-2022}.  Work that is essential for the compilation of reliable editions or the supplementing of important work of literature with annotations. 

The making of children's books has specific features and poses specific questions that have not yet been systematically addressed in genetic-critical studies. The potential of such focus lies in the possibility of a better understanding of how children’s literature is created and what considerations dominate the decision-making process in terms of for example content, character development, construction of setting, or narrative voice and style \cite{Joosen2017}. From the point of view of children's literature and genetic criticism, drafts to \textit{The Brothers Lionheart} are especially interesting due to the novel's seminal position in Lindgren's ouvre, its radical content, the author's well-known difficulties in bringing the novel to its end, and the controversies upon its publication and reception \cite{Andersen2014,Ramstrand2011}. From the perspectives of genetic criticism, HTR, and expert crowdsourcing, the limited but proportionally large amount of shorthand notepads containing drafts to the novel (55 in total) provides a sufficient amount of material/data whilst also enabling a sustainable crowdsourcing life-cycle \cite{dhnb2022}. By mixing the best features of all aforementioned methods, a more coherent, multifaceted whole is created.

\subsubsection{Stylistic Overview of Literary Works} \label{sub:sec:literary-content}

For Lindgren, stenography forged a link between vocalisation and writing which is likely to have favoured oral elements of her work in general \cite{nauwerck2021}. Dialect, folk songs, psalms, and jokes as well as linguistic and onomatopoetic innovations are all recurrent elements of Lindgren's fiction. The LION dataset represent four different works by Lindgren which are written for different purposes and target groups, belong in different genres, and consequently also differ somewhat in style and vocabulary. 

The vocabulary in the excerpt from the text \textit{On our grove} is primarily characterised by the pastoral description of the Swedish grove which it contains, with specific names of flowers, trees, animals, and berries whereas the portion from \textit{Samuel August from Sevedstorp and Hanna in Hult} is slightly more complex in style as it includes allusions, quotes and song lyrics directed toward an older audience, referring for example to historical rural, oral, and religious tradition. The excerpt from \textit{Emil of Lönnerberga} consists of an adaptation of the novel for either screen or stage, and includes song lyrics (\enquote{Bomsicka bom}), directions and dialogue, occasionally written in dialect (for instance a use of \enquote{dä} and \enquote{di} instead of standard Swedish spelling \enquote{det} and \enquote{de}). As for \textit{The Brothers Lionheart}, the vocabulary and style is in line with Lindgren's often expressed credo: that authors of children's books should write in ways that children can easily understand and relate to. A guiding stylistic principle for Lindgren was to use \enquote{common words to say uncommon things} \cite{tornqvist2015}.  Consequently, the vocabulary used by Lindgren in this novel is relatively simple and straightforward. More unusual words include characters (i.e. the dragon \textit{Katla}), fictional place names (i.e. \textit{Nangijala}), and occasionally made-up words and compound words significant both for Lindgren and for the flexibility of the Swedish language. In Melin shorthand, the stenographic representation of compound words may vary depending on for example writeability or space, and exist both as joint and split forms(e.g. duvdrottningen/duv drottningen). 

The phonetic and colloquial spelling of Melin and Lindgren's ideas of how to write for and about children are connected. Many words in narrator and protagonist Skorpan's vocabulary consist of phonetically simplified words, mirroring children's colloquial language. For example word images such as \enquote{huvet} (\enquote{huvudet} -- \enquote{the head}) or \enquote{nitti} (\enquote{nittio} -- \enquote{ninety}), which might be perceived as shorthand abbreviations, but have in fact have been transposed into Lindgren's fiction in the following phase of typing up.
\begin{figure*}[ht]
    \centering
    \includegraphics[width=0.9\textwidth]{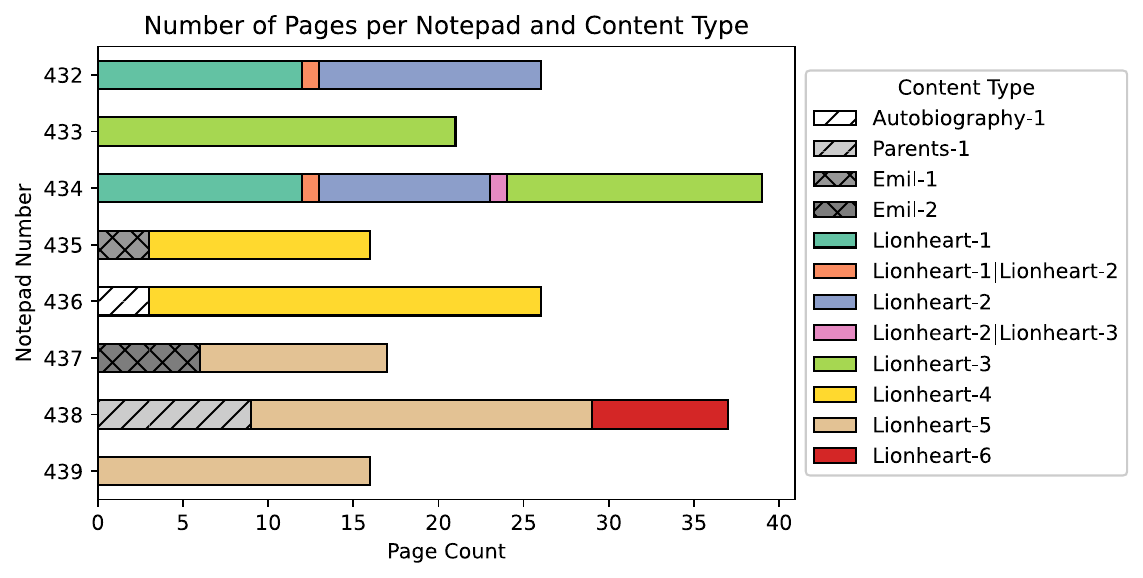}
    \caption{Distribution (in pages) of the different literary works, across the eight notepads.}
    \label{fig:notepad_contents}
\end{figure*}

\subsubsection{Writing and transliterating the Melin system}

The Melin stenographic system in which Lindgren wrote was the standard system taught in Sweden during the 20th century, and has consequently been widespread among secretaries, journalists, and clerical staff. Expert volunteer transliterators of \textit{The Astrid Lindgren Code} are recruited from this group \cite{dhnb2022}.  

Developed by Olof W. Melin in 1890–1892, the Melin system is based partially on the German Gabelsberger system, works according to the frequency of particular sounds in the Swedish language, and uses phonetic symbols to represent vowels, consonants, and consonant combinations, as well as a wide range of abbreviations, prefixes, and suffixes. As is the case with for example Gabelsberger as well as Gregg's and Pitman, stenographers using Melin will deconstruct what they hear, reconstruct it as a sequence of phonetic symbols and shortforms, and finally translate their shorthand notes into typed up longhand. Lindgren's shorthand is on the one hand representative of Melin in terms of following the system's standard closely, but on the other characterised by a tendency to  \enquote{spell out} phonemes of shorthand rather than abbreviating them or relying on, for the system, more advanced shortforms. An obscuring factor is how Lindgren's \enquote{sloppy} handwriting often affects the proportions of scale and slant, which in Melin shorthand are central in producing distinction and meaning.

\subsection{Quantitative Overview}\label{sec:dataset_overview}
\begin{figure*}[ht]
    \centering
    \includegraphics[width=\textwidth]{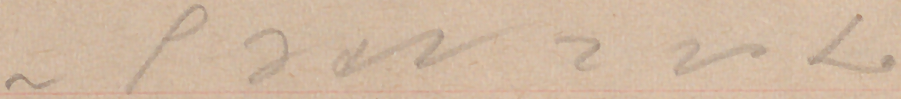}
    \includegraphics[width=\textwidth]{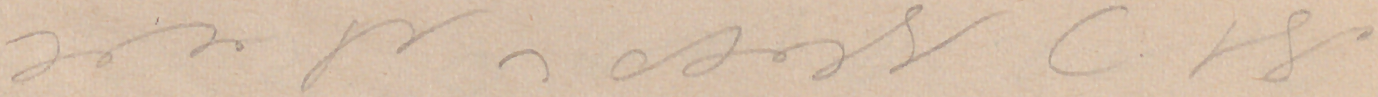}
    \caption{Examples for clean lines.}
    \label{fig:clean_examples}
\end{figure*}

The presented dataset consists of 198 digitised pages, from eight notepads of the manuscript collection of Astrid Lindgren \cite{kb-ref}. The originals, which are part of a collection of 670 notepads, are held at the \textit{Astrid Lindgren Archives} at the \textit{Swedish National Library}, and are being made available through the project \textit{The Astrid Lindgren Code} \cite{al-code}. As outlined above, the eight notepads of \textit{LION} contain excerpts from drafts to four of Lindgren's literary works. We have assigned  the following shorthand titles to each of these works, for ease of referencing: \textit{On our grove}, indicated as \enquote{Autobiography-1}; \textit{Samuel August from Sevedstorp and Hanna in Hult}, indicated as \enquote{Parents-1}; \textit{Emil of Lönneberga}, indicated as \enquote{Emil-1} and \enquote{Emil-2}; and \textit{The Brothers Lionheart}, indicated as \enquote{Lionheart-[1-6]}. 

\autoref{fig:notepad_contents} presents an overview of the content distribution in pages, across the eight notepads. In the case of \textit{The Brothers Lionheart}, portions of the first six chapters, indicated in the figure by their respective number, are contained. Notepads 432 and 434 cover sections of the first two, respectively, three, chapters, including three transitional pages, where the previous chapter ends and the subsequent one continues within the same page, as indicated by \enquote{Lionheart-1\textbar{}Lionheart-2}, respectively \enquote{Lionheart-2\textbar{}Lionheart-3}. Regarding the content stemming from \textit{The Brothers Lionheart}, it should also be noted that for chapters one, two and three, two versions each are contained in the dataset. 

\begin{figure}[b]
    \centering
    \includegraphics[width=\linewidth]{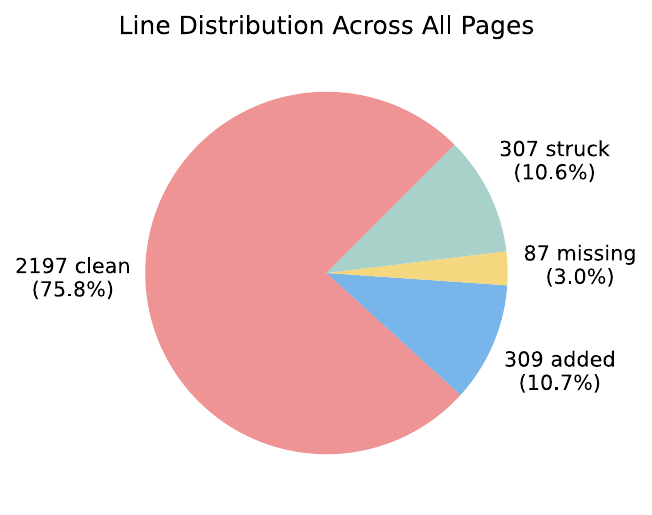}
    \caption{Distribution of lines into the four different categories -- clean, struck, added and missing.}
    \label{fig:line_pie}
\end{figure}

For the other three works, the notepads only contain short excerpts, spanning a few pages each. In contrast to \textit{Lionheart-[1-6]}, the numbering here does not indicate a chapter relation but is simply used as a counter, and, in the case of \textit{Emil}, to differentiate between two portions of text, written in two different notebooks (i.e. 435 and 437). Lastly, it should be noted that some of the presented notepads originally contain private notes and letters by the author. For privacy reasons, these have been removed and are not shared or considered in this work.

\begin{figure*}[ht]
    \centering
    \includegraphics[width=\textwidth]{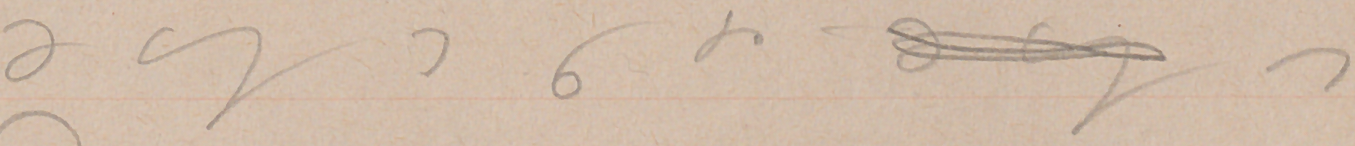}
    \includegraphics[width=\textwidth]{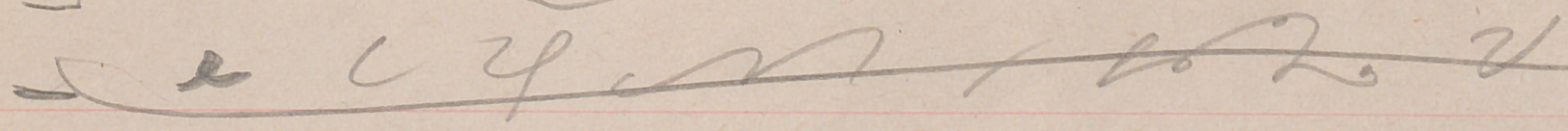}
    \caption{Examples for lines containing varying amounts and styles of strikethrough.}
    \label{fig:strikethrough_examples}
\end{figure*}

\begin{figure*}[ht]
    \centering
    \includegraphics[width=\textwidth]{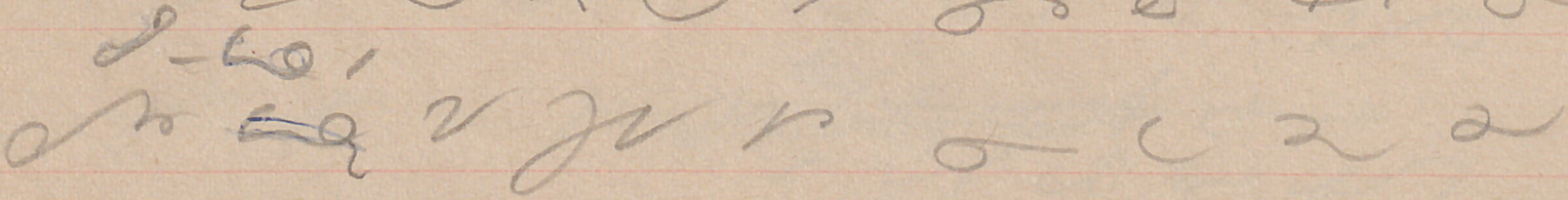}
    \includegraphics[width=\textwidth]{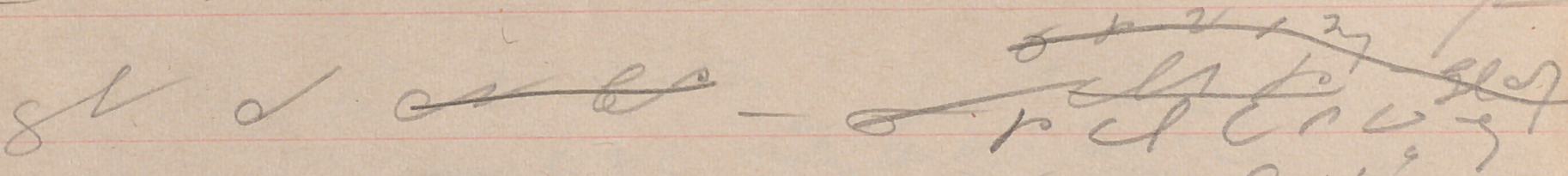}
    \caption{Examples for lines containing individual additions (top), and in combination with struck-through words (bottom).}
    \label{fig:additions_example}
\end{figure*}

All of the pages have been segmented into handwritten lines (cf. \autoref{sec:data_prepr} \nameref{sec:data_prepr}), resulting in 2900 separate images that can for example be used for handwritten text recognition and document image processing. Several of the lines bear a variety of editorial marks, summarised in \autoref{fig:line_pie}. Concretely, about 10\% of the lines contain at least one word that has been struck-through, indicating the author's intention to delete it. Two examples for varying degrees of strikethrough, and their impact on readability, are shown in \autoref{fig:strikethrough_examples}. Another $10\%$ of the lines entail additions, i.e. words written above, and occasionally below, the already written line, indicating corrections or additions. The latter $10\%$ often also contain struck-through portions, where one or more words are replaced by an addition. However, we do not differentiate further in this regard and combine all of these in the general \textit{additions} category. Examples for differing amounts of additions, and combinations with strikethrough, are shown in \autoref{fig:additions_example}. 

Lastly, as presented in \autoref{fig:line_pie}, the dataset contains a further $3\%$ of lines, which are indicated as missing. For these, the transliterations are incomplete, for example due to severe obfuscations by strikethrough strokes. These 87 lines are included in the data repository, alongside their partial transliterations. However, we do not consider them further in this work and exclude them from all experiments presented below. The 2197 lines that do not feature any of the characteristics above, are denoted as \textit{clean}. \autoref{fig:clean_examples} shows two examples of such lines.

\subsection{Data Preparation}\label{sec:data_prepr}
The preparation of the dataset entails the collection of transliterations, the acquisition and line-level segmentation of the archival images, and the combination of the two pieces of information to obtain the final, annotated data. Each of these steps is briefly summarised below. 

\subsubsection{Transliterations}
The transliterations were provided by trained stenographers, via expert crowdsourcing in a peer editing process \cite{dhnb2022}. It should be noted here that we use the term \enquote{transliteration}, instead of the commonly used \enquote{transcription}, as the former describes the representation of one alphabet in another, which is the case when converting stenography to Latin characters. During the transliteration process, the stenographers indicated struck-through words and additions, both of which were converted to a suitable dataformat afterwards. Besides this, the experts indicated line breaks in the transliterations, corresponding to the line endings in the page images. 

\subsubsection{Image Acquisition and Segmentation}
The page images were digitised by the \textit{Swedish National Library}, using a copy stand (i.e. top-down) camera setup. All eight notepads follow the same general layout, an example of which is shown in \autoref{fig:notepad_sample}. The sheets of toned paper are bound together by a spiral binding. Each page contains ruling, in the form of 15 red, printed lines that are evenly spaced, except for a larger margin towards the top and bottom of the page. These landmarks, together with the distinct edges of the notepad against the digitisation background, were used to perform a preliminary segmentation of the page and its individual lines. 

The majority of the words were written in lead pencil, with the exception of a few sections where Lindgren used a blue ballpoint pen. Word segmentation proposals were obtained using a combination of thresholding, morphological operations and connected component labelling. The word bounding boxes and their assignment to a segmented line, were manually refined and proofread.

\subsubsection{Alignment}
As a final step in the dataset preparation process, line-level annotations were obtained by combining the segmented lines with the transliterated text-lines. We did not perform further manual alignment steps, to connect word-level transliterations with their corresponding bounding boxes, due to limited proofreading capacities. However, the bounding box coordinates are included in the data repository. In the case of clean and struck lines, word-level annotations of utilisable confidence may be obtained by sequentially combining the transliterations and bounding boxes within a given line. Due to the challenging reading-order of lines, containing additions, such an automated approach is expected to fail for this specific line type. 



\begin{table}[hbt]
    \centering
        \caption{Line count per datasplit and line type.}
    \label{tab:dataplit_summary}
    \begin{tabular}{c|c|c|c||c}
        Datasplit & Clean & Struck & Added & Split Total\\
        \hline
        Train & 1224 & 196 & 200 & 1620\\
        Validation & 306 & 49 & 50 & 405\\
        Test-LH & 474 & 31 & 24 & 529\\
        Test-OOV & 191 & 31 & 34 & 256\\
        \hline
        \hline
        Type Total & 2195 & 307 & 308 & 2810\\
    \end{tabular}
\end{table}

\subsection{Data Splitting Considerations}
Besides providing the raw data, consisting of page and line images and corresponding transliterations, we propose a number of datasplits that can be used for various deep learning tasks. We designate a portion of the data to be used during training and hyperparameter fine-tuning, using five-fold cross validation. The remainder of the data is set aside as test set for the final evaluation. 
Considering the unbalanced distribution of content types (cf. \autoref{sec:dataset_overview}), the datasplits have been arranged in a way to allow the investigation of model performances on the majority content type, \textit{The Brothers Lionheart}, and the generalisation to the other included literary works by Astrid Lindgen. 

Based on these considerations, we designate all lines belonging to chapter four of \textit{The Brothers Lionheart} (i.e. Lionheart-4) as the in-vocabulary test set, referred to as \enquote{Test-LH}. All lines belonging to Autobiography-1, Parents-1, Emil-1 and Emil-2 are set aside as out-of-vocabulary test set,  \enquote{Test-OOV}.

The proposed datasplits are summarised in \autoref{tab:dataplit_summary}. The lines within each datasplit can either be considered as a whole, i.e. combining clean, struck and added lines, or in various subset combinations, such as only clean lines. For convenience and reproducibility, corresponding lists for all of these combinations are included in the data repository.

\begin{figure}[htb]
    \centering
    \includegraphics[width=0.9\linewidth]{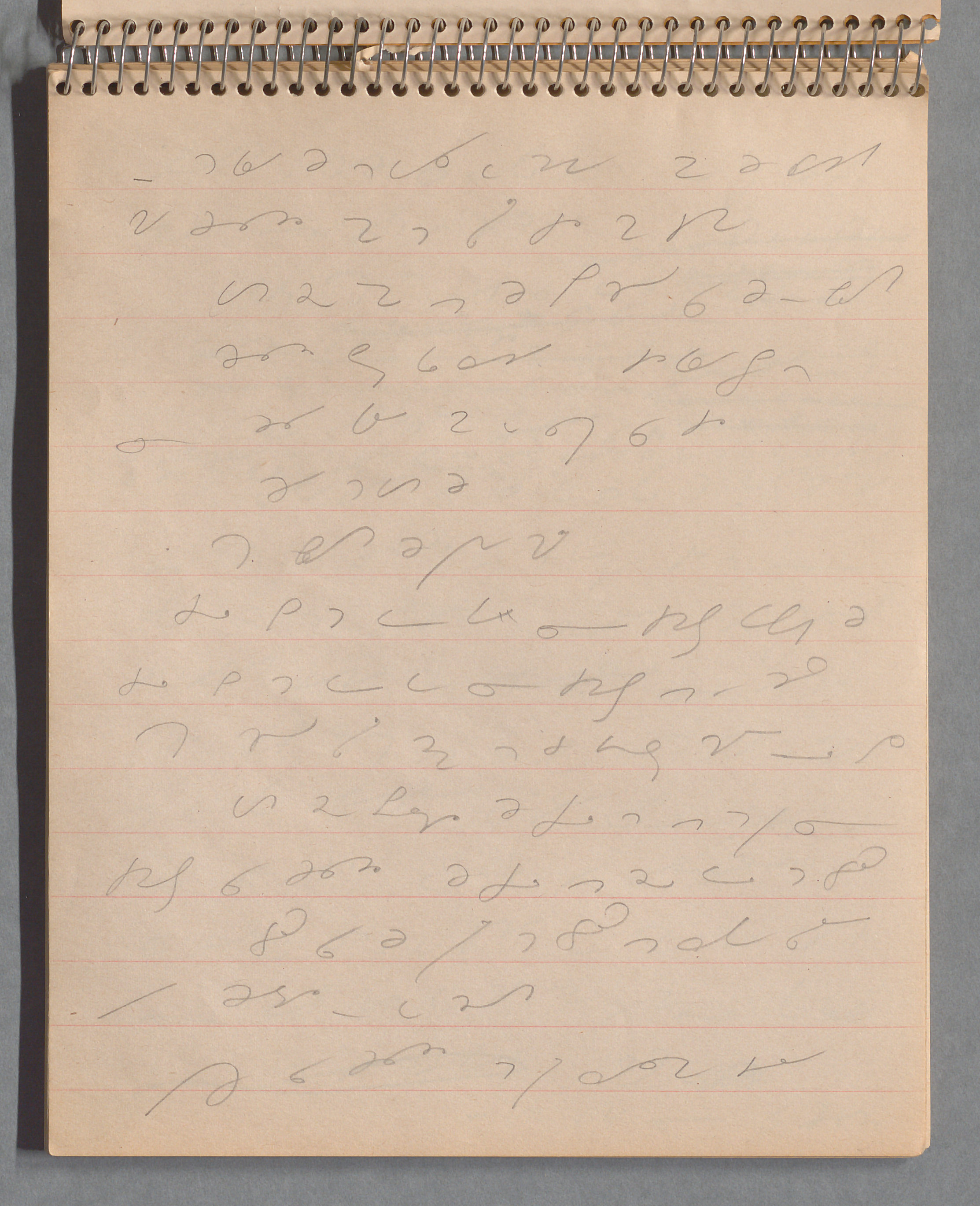}
    \caption{Sample page, demonstrating the original contrast, metal binding at the top, and red printed rulings.}
    \label{fig:notepad_sample}
\end{figure}

\subsection{Brief Quantitative Analysis of Textual Content}

\begin{figure}[hbt]
    \centering
    \includegraphics[width=\linewidth]{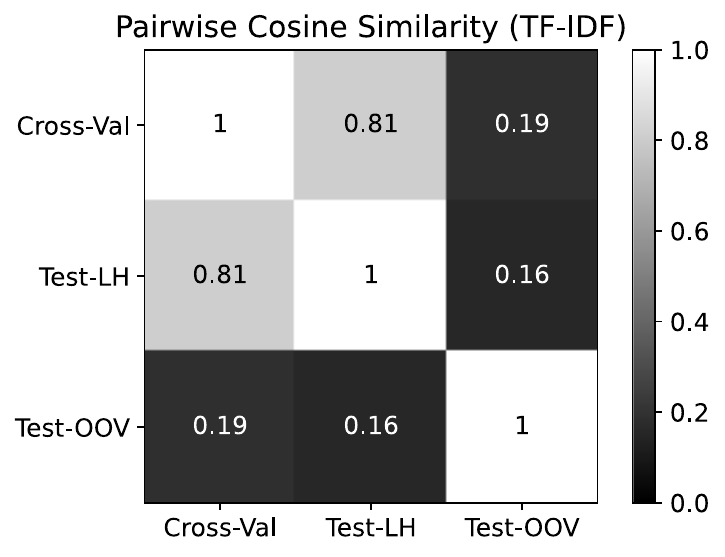}
    \caption{Cosine similarity between the TF-IDF vectors (excluding stop words) of the respective datasplits.}
    \label{fig:cosine-sim}
\end{figure}

In order to quantitatively summarise the textual contents of the cross-validation and the two test sets (Test-LH, Test-OOV), we perform a brief linguistic analysis of the three sets of documents. For this, we firstly remove all stop words, using the list provided by NLTK \cite{nltk} for Swedish. Afterwards, we calculate the Term Frequency - Inverse Document Frequency (TF-IDF) \cite{tf-idf} scores for the remaining words in the three partitions. Each text can then be represented as a vector of document-specific TF-IDF scores. Calculating the pairwise cosine similarity quantifies the similarity between the respective documents. \autoref{fig:cosine-sim} summarises these pairwise scores. As can be seen from the figure, the cross-validation and Test-LH portions share a considerable overlap, whereas Test-OOV is noticeably different from both. This matches the aforementioned content descriptions, with the two former documents stemming from the same corpus, and the latter being a combination of three distinct other corpora.

Taking the list of the ten words with the highest TF-IDF scores per document, presented in \autoref{tab:top10tfidf}, into consideration, the observations above are further emphasised. The lists for the cross-validation split and Test-LH feature figures (\enquote{Jonatan}, \enquote{Sofia}) and place names (\enquote{Nangijala}, \enquote{Körsbärsdalen}) that are central to \textit{The Brothers Lionheart}. In contrast to this, the list for Test-OOV contains a combination of central figures and places from \textit{Emil of Lönneberga} (Emil and his parents -- \enquote{mamma}, \enquote{pappa} -- and the tool shed - \enquote{snickerboa}), and the names of Lindgren's parents (\enquote{Hanna} and \enquote{Samuel August}). The one word that occurs in all three lists, \enquote{mej}, exemplifies the use of colloquial spellings, mentioned initially. The standard spelling for this word is \enquote{mig} (English: me, myself). In this regard, it should be noted, that we do not control for alternative spellings in our initial filtering of stop words. Therefore, \enquote{mej}, which is considered a stop word in its official spelling, still appears in this TF-IDF-based word list.

\section{Baseline: Handwritten Stenography Recognition}\label{sec:clean-baseline}

In order to establish a baseline for handwritten stenography recognition on the LION database, we train and evaluate a state-of-the-art HTR model on the set of clean lines. 

\subsection{Model: Gated-CNN-BGRU}
\begin{figure}[ht]
    \centering
    \includegraphics[width=\linewidth]{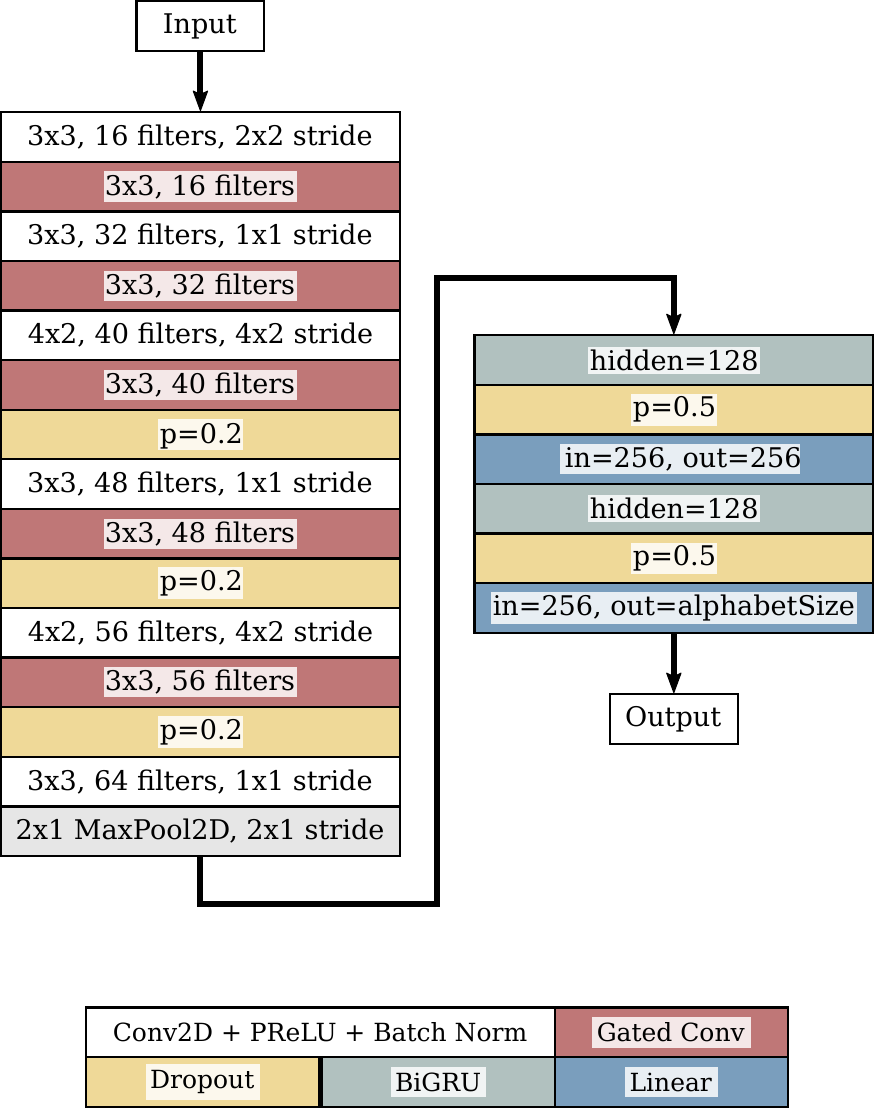}
    \caption{Summary of the Gated-CNN-BGRU architecture as used in the presented experiments.}
    \label{fig:architecture}
\end{figure}

All of the experiments presented in this paper are performed using a slightly modified version of the Gated-CNN-BGRU architecture, proposed by Neto et al. \cite{flor_restricted}. This model has been shown to perform well in limited-resource settings of similar extent as the LION database. Furthermore, this architecture outperformed other CTC-based approaches in our initial experiments. The architecture consists of two major components, shown in detail in  \autoref{fig:architecture}, that are trained in an end-to-end fashion. Based on prior experiments, we replace the originally proposed Batch Renormalisation \cite{br} layers with regular Batch Normalisations \cite{batch_norm}. Furthermore, unlike Neto et al. \cite{flor_restricted}, we employ best path decoding \cite{ctc_graves}, instead of a language model, to obtain the transliterations, in order to focus on the performance of the text recognition approach.

\begin{table}[hbt]
    \centering
        \caption{Top-10 terms, weighted by TF-IDF score, for each of the datasplits.}
    \label{tab:top10tfidf}
    \begin{tabular}{c|c}
        Split & Top-10 Terms \\
        \hline
        Cross-Val & Jonatan, sa, mej, kom, bara \\
        & Sofia, Nangijala, stod, också, sen\\
 \hline
 Test-LH & Jonatan, sa, Sofia, mej, Körsbärsdalen\\
 & ja, just, kom, sen, red\\
 \hline
 Test-OOV & Emils, Hanna, hopp, snickerboa, Samuel \\
    & August, mamma, mej, fallera, pappa \\
 \end{tabular}

\end{table}

\subsection{General Training and Evaluation Protocol}\label{sub:protocol}
The baseline model is trained for up to 100 epochs, using the AdamW \cite{adamw} optimiser with a learning rate of 0.001, a batch size of eight and the standard CTC loss. All line images are preprocessed by first converting them to the HSV colour space and obtaining a single-channel image by discarding the hue and saturation channels. This step was performed instead of a regular greyscale conversion as it removes most of the printed red rulings (cf. \autoref{fig:notepad_sample}). Afterwards, the remaining value channel is inverted and the contrast is stretched, using the second and 98th intensity percentiles as boundaries. 

During training, the validation CTC loss is measured after each epoch. Following an initial warm-up period of ten epochs, early stopping with a patience of ten epochs is applied, based on the validation loss. The model weights of the best-performing validation epoch are preserved for the final evaluation. In order to better ascertain the variability of the model performance, training is repeated from scratch 30 times per fold, yielding a total of 150 sets of weights, and thus results. All trained models are evaluated on the test sets, measuring the Character Error Rate (CER) and Word Error Rate (WER), which are defined as follows: 

\begin{equation}
ER = \frac{S + D + I}{N}
\end{equation}

where $S$, $D$ and $I$ are the number of character (word) substitutions, deletions and insertions, respectively, that need to be performed to convert a given text to a reference text. The sum of these transformations corresponds to the Levensthein distance \cite{levenshtein}. $N$ indicates the amount of characters (words) in the reference string. For both metrics, lower values are better, with $0$ being the optimum.

\subsection{Experiment}
As mentioned above, we limit our experiments on the new LION dataset to the portion of clean lines. Besides excluding lines with any form of strikethrough or additions during training and validation, we also exclude these lines when reporting the model's performance on the test set. We chose to limit the data for this first investigation, in order to rule out any potential side-effects of the altered lines. Furthermore, many of the currently available datasets either do not contain such forms of alterations (e.g. Saint Gall \cite{saint-gall}), or explicitly exclude them, for example by providing placeholder transcriptions, like \enquote{\#}, for struck-through words (e.g. IAM \cite{iam}). The recognition performance for struck and added lines is briefly discussed in \autoref{sec:struck_added_results}.

\subsection{Baseline Results and Analysis}


\autoref{tab:clean-results} shows the mean CER and WER of the baseline experiment. As shown in the table, there is a noticeable difference in performance between Test-LH and Test-OOV, with the model performing considerably worse on the latter portion. Regarding the overall performance (third row), it can be noted that both error rates are noticeably higher than the rates obtained on commonly-used benchmark datasets of similar size, such as Saint Gall \cite{saint-gall}, for which for example Neto et al. achieve an error rate around 4\% \cite{flor_restricted}. The comparably high error rates illustrate the challenging nature of the LION dataset. Additionally, the reduced performance on the out-of-vocabulary test set (Test-OOV) gives an indication of the difficulty of stenography recognition itself. 


\begin{table}[t]
    \centering
    \caption{CER and WER (in \%) for the baseline experiment, using the Gated-CNN-BGRU, trained on the original, clean lines.}
    \label{tab:clean-results}
    \begin{tabular}{c|c|c}
         Data Split & CER (Std. Dev.) & WER (Std. Dev.)\\
        \hline
        Test-LH & 26.88 (0.97) & 51.92 (1.35)\\
        Test-OOV & 38.70 (0.91) & 65.19 (1.25)\\
        \hline
        Combined & 29.81 (0.92) & 55.14 (1.26)\\
    \end{tabular}
\end{table}

\section{Encoding Stenographic Domain Knowledge}\label{sec:encoding}

The baseline experiment, presented in the previous section, treats the stenography recognition as a traditional transcription problem. So far, we have not considered any of the aspects that are inherent to stenography and the Melin system, and that set it apart from other scripts, such as Latin. One major characteristic of this stenographic system is that the symbol set is considerably larger than that of the Swedish alphabet, resulting in a one-to-many mapping of symbols to characters in the Swedish transliterations, and thus also in the CTC decoding step. In order to investigate whether more direct mappings, closer to a one-to-one relationship, can improve the recognition performance, we have selected four groups of such mappings and implement target sequence \textit{encodings}, inspired by diplomatic transcriptions (cf. e.g. \cite{diplomatic}). 

\subsection{Encoding Schemes}

Firstly, we consider words that share the same visual representation as a character symbol, for example \enquote{och} (\textit{and}), being written as the symbol for \enquote{o}, \textit{jag} (\textit{I}) as \enquote{j} and \enquote{var} (\textit{where}, \textit{was}) as \enquote{v}. In total, we have identified 14 such \textit{shortforms} and their corresponding characters, which are shown in \autoref{tab:shortform-list}. The second row in \autoref{tab:encodings} shows an example of applying this encoding technique, which we term \textit{shortform}. Although shortforms may also be used when the word appears as a prefix (e.g. \textbf{över}allt), we limit the encoding to isolated occurrences of the respective words. The decoding of prefix occurrences is ambiguous (e.g. \textbf{ö}gon vs \textbf{ö}[ver]allt) and would therefore require additional language knowledge to definitively decode a given string. 

For the second encoding scheme, referred to as \textit{suffix}, we selected four frequently appearing suffixes, \enquote{-are}, \enquote{-ing}, \enquote{-en} and \enquote{-et}, and replace each of the occurrences with its own symbol, as demonstrated in the third row of \autoref{tab:encodings}. The first two prefixes were primarily chosen because they are represented by their own symbols in the Melin standard. In addition to this, the latter two are included because they are often indicated by leaving out the \enquote{e} and simply appending an \enquote{n}, respectively \enquote{t} as terminating character. Although this cannot be considered a separate symbol of its own, it does result in a one-to-many character mapping like the other explored encodings.

As a third encoding method, termed \textit{n-gram}, we investigate the 31 n-grams, shown in \autoref{tab:ngram-list}, for which the Melin system defines its own symbols. We emulate this symbol assignment by replacing the respective n-gram occurrences with separate symbols in the target sequences. An example for this is shown in row four of \autoref{tab:encodings}, where the encoding of the n-gram \enquote{nkt} is visualised as \enquote{\&}. 

Lastly, we combine a variety of (sub-) words for which the Melin system defines its own symbols, creating a more extensive set of transformations, as compared to the aforementioned three, therefore termed \textit{Melin}. Concretely, we consider commonly-used shortforms, as well as words that are represented by their own symbols. Besides this, we consider a selection of affixes and n-grams for which the Melin standard defines individual symbols. \autoref{tab:melin} summarises the considered words and n-grams, and the final row in \autoref{tab:encodings} demonstrates the application of this encoding. 

Overall, it should be noted that these four encoding schemes were chosen and implemented in a manner that makes them fully reversible, i.e. all altered strings can be unambiguously decoded to their original representation by replacing the respective placeholder symbols with the characters they represent. A variety of other encodings schemes are conceivable within the Melin standard, however these are often not unambiguously decodable, or would require the integration of additional language knowledge into the decoding process. 


\begin{table}[t]
    \centering
        \caption{Demonstration of applying the four different encoding schemes on a sample string.}
    \label{tab:encodings}
    \begin{tabular}{c|llllll}
         Original & jag & tänkte &att& det& var& finare\\
         \hline
         Shortform & \textbf{j}& tänkte &att& det &\textbf{v} &finare\\
         \hline
         Suffix  &  jag &tänkte &att &d\textbf{*} &var& fin$\boldsymbol{\alpha}$\\
         \hline
         N-gram  &  jag& tä\textbf{\&}e &att& det& var& finare\\
         \hline
         Melin  & $\boldsymbol{\beta}$ &$\boldsymbol{\gamma}$te &att & $\boldsymbol{\delta}$& $\boldsymbol{\epsilon}$ &fin$\boldsymbol{\alpha}$\\
    \end{tabular}
\end{table}

\begin{table}[t]
    \centering
    \caption{List of shortforms and the character by which they are replaced during the \textit{shortform} encoding step.}
    \label{tab:shortform-list}
    \begin{tabular}{ccccc}
         av $\rightarrow$ a & bar $\rightarrow$ b & de $\rightarrow$ d & en $\rightarrow$ e & från $\rightarrow$ f \\
         har $\rightarrow$ h& jag $\rightarrow$ j &kan $\rightarrow$ k&men $\rightarrow$ m & och $\rightarrow$ o\\
         ut $\rightarrow$ u & var $\rightarrow$ v & är $\rightarrow$ ä & över $\rightarrow$ ö
    \end{tabular}
\end{table}

\begin{table}[t]
    \centering
    \caption{List of n-grams considered for the \textit{n-gram} encoding scheme. N-grams are resolved from biggest to smallest, to avoid conflicts due to overlaps, e.g. "nskt" vs "skt".}
    \label{tab:ngram-list}
    \begin{tabular}{cccccccc}
         nsion& nskt& nsj& nst& nsk& nkt& skt& ng\\
         sj&         tj&br& fr& tr& kv& tv& sk\\
         sl& sm & sn& sp & st& sv& nt& nd\\
         ns& nj& nk &          bt& pt& kt &          xt \\
    \end{tabular}
\end{table}

\begin{table}[ht]
    \centering
    \caption{Summary of the \textit{Melin} encoding scheme. Groups from top to bottom: shortforms, prefixes, suffixes, n-grams.}
    \begin{tabular}{ccccccc}
    alldeles& bland& bättre& de& den& det \\
    då & där& efter& eller& en& ett \\
    genom& gång & har& honom& här& ingen \\
    inte& jag& just & kan& kom& kunna\\
    med& men& mot& mycket& måste & och\\
    om& själv& skulle& som & under &  upp \\
    var & varit& är& även\\
    \hline
    an-& be-& fort-& fram-& hän-& in- \\
    kon-& kun-& kän-& lång-& märk-& någo-\\
    på-& slut-& särskil-& till-& tänk-& ut-\\
    verk-& vill-& över-\\
    \hline
    -ande& -de& -are& -het& -kring& -ning\\
    -ing\\
    \hline
    -ng-& -sj-& -tj-& -br-& -fr-& -tr-\\
    -kv-& -tv-& -sk-& -sl-& -sm-& -sn-\\
    -sp-& -st-& -sv-& -nst-& -ns-& -nk-\\
    -nkt-& -av-& -för-\\
    \end{tabular}
    \label{tab:melin}
\end{table}

\subsection{Experiments}
We use each of the encoding schemes, introduced above, in order to create alternative target text representations. These encoded texts are then used to train four versions of the Gated-CNN-BGRU, using the training protocol introduced above (\autoref{sub:protocol}). We adapt the alphabet size for each of the experiments according to the respective encoding, for example increasing it by 14 symbols for the \textit{shortform} encoding. Before calculating the CER and WER on the test set, the predicted text lines are decoded back to the regular Swedish alphabet, i.e. inverting the encoding step and obtaining a regular Swedish text representation.

\subsection{Results and Analysis of Different Encoding Schemes}
\autoref{tab:base_encoding_results} summarise the results for the four encoding methods. Generally, the error rates lie within $\pm0.3$ percentage points (pp) of the baseline performance. A notable exception is the WER obtained by the \textit{Melin} encoding scheme, which yields a considerable improvement of 3pp. A potential explanation for this improvement lies in the shortform portion of the encoding scheme (first group in \autoref{tab:melin}). An inspection of the (mis-) spellings of these words reveals that roughly 74\% of these words are spelt correctly when using the \textit{Melin} encoding scheme, an increase of approximately 13\% over the baseline recognition of the same words. 

Considering the overall frequency of words that are transformed by the different encoding schemes, it can be noted that some of the considered symbols only appear in very small numbers, for example a few tens for some of the n-grams, out of the almost 10\,000 words. This low number of samples may be a contributing factor to the small impact of the different encoding schemes. In an attempt to mitigate this, we therefore study the recognition performances further, using a larger, synthetic training set in the last set of experiments, discussed in the following section.

\begin{table}[ht]
    \centering
    \caption{Summary of CER and WER (in \%) for the different encoding schemes and the encoding-free baseline, trained and evaluated on the original, clean lines.}
    \label{tab:base_encoding_results}
    \begin{tabular}{c|c|c}
         Encoding & CER (Std.Dev.) & WER (Std.Dev.)  \\
         \hline
         None & 29.81 (0.92) & 55.14 (1.26) \\
         Shortform & 29.75 (0.92) & 54.81 (1.33)\\
         Suffix & 29.78 (1.16) & 54.81 (1.42)\\
         N-gram & 29.86 (1.43) & 54.21 (1.76)\\
         Melin & 29.97 (1.36) & 51.99 (1.53)\\
    \end{tabular}
\end{table}

\section{Training on Synthetic Data}\label{sec:pretrain}
As outlined above, we expand the training set, in order to better study the potential of our proposed encoding schemes. A commonly-used approach for increasing the training set size is to use data augmentation techniques.
We studied this approach extensively in our prior work and only obtained small, albeit significant improvements \cite{ibpria}. In the following, we explore an alternative approach, based on the recombination of individual words. 

\subsection{Dataset Creation}
In order to create a variety of synthetic line images and corresponding transliterations, we firstly segment the original training lines, using the bounding boxes provided with the dataset. Word labels are obtained by aligning the segmented word images with the provided transliterations. As mentioned initially, this level of annotations has not been proofread. We therefore discard all lines, a total of 20, for which the number of bounding boxes differs from the number of words in the transliteration. While the pool of remaining words is not guaranteed to be perfectly annotated, most of the alignments can be expected to be correct, due to prior proofreading efforts. 

Based on the obtained pool of annotated word images, new text lines are created via random combinations. Line-breaks are inserted such that the resulting text lines have a similar character count as the original ones. Before pasting the word images onto a blank canvas, slight transformations are applied, using the previously identified augmentations: rotation, scaling and shifting \cite{ibpria}. Overall, each original word image is used ten times in different word contexts to generate new lines. Regarding the content of the newly generated lines it should be noted that these are combined in an unconstrained manner, not taking any linguistic considerations into account. 

We apply this generation process to the training sets of the five folds, yielding five sets of larger datasets, each containing around 9\,400 line images. The validation and test sets are not altered by the generation process and remain in their original configuration.

\subsection{Experiments}
The training and evaluation follow the same protocol as above, with the exception of replacing the training sets with the newly generated, synthetic ones for each of the folds. Following the initial pre-training stage on the synthetic data, a fine-tuning step is applied, using the genuine, original training set of the respective fold. The fine-tuning stage follows the same parameters as all other experiments, with the exception of removing the warm-up period, as the models are not being trained from scratch, as is the case for all other experiments.

Where applicable, statistical hypothesis testing is performed based on the mean line CER, respectively WER, using Wilcoxon paired signed-rank tests \cite{wilcoxon} with a Bonferroni correction \cite{bonferroni} to correct for multiplicity.

\subsection{Results and Analysis}

The results of the combined pre-traning and fine-tuning step are summarised in \autoref{tab:pretrain_results}. Comparing the encoding-free performance with the baseline (\autoref{tab:clean-results}), clear improvements for both metrics can be observed. Similarly, each of the examined encoding schemes outperforms its baseline counterpart (\autoref{tab:base_encoding_results}) by $3.8-5.3$pp for the CER and $6.4-8.3$ pp for the WER. 

The three encoding schemes \textit{shortform}, \textit{suffix} and \textit{Melin} significantly ($p<0.01$) outperform the encoding-free setup, when applied in conjunction with pre-training and fine-tuning. Similar to the original encoding experiments, a considerable improvement of 3.3 pp is achieved for the WER by the \textit{Melin} scheme. In a traditional recognition setup, decreases in the WER are often coupled with considerably larger decreases for the CER, due to a difference in scaling (character vs. word count). However, for the \textit{Melin} scheme, the CER improvement is more modest, compared to the one for the WER. One explanation for this lies in the considered shortforms, which amount to almost 30\% of the words in the test sets. All of these words are represented by a single symbol, each. A correctly recognised shortform symbol will therefore result in a correctly recognised word, assuming that the spaces alongside it are also correctly identified. This will positively impact both metrics. At the same time, the incorrect recognition of a regular character as a shortform (or vice-versa) will have a much larger adverse effect on the CER than a regular character-character confusion, as illustrated in the example in \autoref{tab:confusion_example}. This effect also applies to the \textit{shortform} encoding, however, as this only entails the 14 character shortforms, which amount to roughly 15\% of the test words, the impact is not as drastic, as for the \textit{Melin} scheme. 

Lastly, considering the \textit{n-gram} encoding, the obtained results are worse, respectively differ only marginally from the encoding-free version. One conceivable explanation is that the encoding was applied in a naive fashion, replacing any occurrences of a given n-gram with its respective symbol. This may result in substitutions that are not in line with the Melin system, for example, when an n-gram occurs as the result of a compund word, such as \enquote{ns} in \enquote{sammansättning} (samman + sättning). In order to investigate this further, the integration of language knowledge, possibly on a phonetic level, or more fine-grained stenography annotations, e.g. at symbol level, will be required. 

Overall, three of the four examined encoding schemes have significantly improved the recognition performance, demonstrating the positive impact of carefully integrating stenographic knowledge into the text recognition process. At the same time, the positive impact of pre-training on a larger, synthetic dataset emphasises the challenge of low-resource datasets. 
\begin{table}[ht]
    \centering
    \caption{Summary of CER and WER (in \%) for the different pre-trained and fine-tuned encoding schemes and the encoding-free baseline.}
    \label{tab:pretrain_results}
    \begin{tabular}{c|c|c}
         Encoding & CER (Std.Dev.) & WER (Std.Dev.)  \\
         \hline
         None & 25.68 (4.38) & 48.15 (4.78)\\
         Shortform & 24.49 (0.83) & 46.55 (1.45)\\
         Suffix & 25.00 (0.81) & 47.13 (1.37)\\
         N-gram & 26.04 (1.84) & 47.79 (2.46)\\
         Melin & 24.65 (0.86) & 44.82 (1.45)\\
    \end{tabular}
\end{table}

\begin{table}[ht]
    \centering
    \caption{Example, demonstrating the impact of character-character confusion compared to character-shortform confusion.}
    \label{tab:confusion_example}
    \begin{tabular}{llc}
       Ground truth  & alla \\
       Char-char confusion & all\textbf{b} & 25\% CER\\ 
       Char-shortform confusion & all\textbf{alldeles} & 175\% CER \\
    \end{tabular}
\end{table}

\begin{table}[ht]
    \centering
    \caption{Summary of CER and WER (in \%) for the three line types, \textit{struck}, \textit{added} and \textit{clean}, based on the pre-trained and fine-tuned encoding-free model.}
    \label{tab:altered_lines}
    \begin{tabular}{c|c|c}
        line type & CER (Std.dev.) & WER (Std.dev.)\\ 
        \hline
        struck & 56.48 (2.47) & 78.00 (2.25)\\ 
        added & 68.44 (2.01) & 90.15 (1.22)\\
        \hline
        clean & 25.68 (4.38) & 48.15 (4.78)\\
    \end{tabular}
\end{table}

\section{Recognition of Struck and Added Lines}\label{sec:struck_added_results}

A detailed study of the recognition performance for struck and added lines, as well as the impact of their presence during training, is beyond the scope of this paper. However, to provide a general performance overview of this considerable portion of the dataset (around 20\% of lines), \autoref{tab:altered_lines} presents the CER and WER for the struck and added lines of the test set, using the pre-trained and fine-tuned model, without any encoding applied. For convenience, the error rates for the clean lines are repeated in the final row.

The recognition performance is considerably lower for both types of lines than that for the clean lines. Considering the struck lines, one option for improving the performance can lie in the removal of strikethrough strokes, as for example proposed in \cite{das_strike,poddar}. In addition to this, including a substantial amount of struck lines in the training set may enable the model to recognise struck words regardless of the occlusion. 

In the case of added lines, we do not expect the latter approach to work, as a large portion of the recognition issues stem from the structure of the chosen model, which outputs one character per time-step, without any spatial information. Referring back to \autoref{fig:additions_example} (top), it can be observed that in the first quarter of the line, two characters need to be recognised per timestep, including the information which of the transliterated lines a recognised character belongs to. Approaches for recognising paragraphs or whole pages have for example been proposed by Yousef and Bishop \cite{origami}, and Bluche \cite{bluche-paragraph} and could be an interesting starting point for future work in this regard.

\section{Conclusion}\label{sec:conclusions}



In this work we have established a baseline for handwritten text recognition for stenography, using the newly introduced LION dataset. We have studied the effect of integrating stenography-based domain knowledge, in the form of selected encoding schemes, derived from visual aspects and rules of the \textit{Melin} writing system. In addition to this, we have investigated the use of pre-training and fine-tuning, using generated line images, to increase the volume of the, otherwise low-resource, LION dataset. Based on our experiments and the obtained results, we draw the following conclusions:

\begin{enumerate}
    \item Automatically transliterating handwritten stenography poses a challenging task to handwritten text recognition.
    \item Pre-training on generated line images, followed by a fine-tuning step using genuine data, improves the recognition performance considerably.
    \item Combining pre-training and fine-tuning with selected, stenography-based target sequence encoding schemes yields further, significant recognition improvements.
\end{enumerate}

Overall, it can be noted that the transliteration of stenography is not straightforward, as the writing system consists of an extensive set of rules, unlike many other, previously examined scripts. Despite this, we have demonstrated that handwritten stenography recognition is possible. The produced transliterations can, for example, be used as a basis for human proofreading, thus considerably reducing the time and effort, required to process a document. 

This aspect is especially relevant, as the first-hand, applied knowledge of stenography is steadily decreasing. It is therefore crucial to utilise these skills and experiences while they are still readily available, in order to process as many stenographic documents as possible, ensuring future access to the material. 

Unlocking the stenographic notes of Astrid Lindgren, one of Sweden's most renowned authors, is of significance from a cultural heritage perspective. Beside this, however, handwritten stenography recognition is also of relevance for areas such as political history and genealogy, for example in the form of stenographed court records, respectively personal diaries. 

We therefore make the LION dataset publicly available with the aim of encouraging future research in handwritten stenography recognition. Potential avenues for future work are, for example, approaches that generate entirely new, unseen word images using the Melin system, or that integrate phonetic domain knowledge into the recognition process. Besides this, investigating linguistic approaches, for example in the form of a language model, may be of interest.

\backmatter

\bmhead{Data Availability}\label{sec:data_avail}
The LION dataset is available in the following Zenodo repository: \textit{submission in progress}. Furthermore, the code for the presented experiments can be obtained from this Zenodo repository: \url{https://doi.org/10.5281/zenodo.8249817}.

\bmhead{Acknowledgements}
The authors would like to acknowledge the extensive help provided by volunteers, transliterating and proofreading the stenographic manuscripts, without whom this project would not have been possible.


This work is partially supported by Riksbankens Jubileumsfond (RJ) (Dnr P19-0103:1). The computations were enabled by resources provided by the National Academic Infrastructure for Supercomputing in Sweden (NAISS) at Chalmers Centre for Computational Science and Engineering (C3SE) partially funded by the Swedish Research Council through grant agreement no. 2022-06725. 

\bibliography{main.bbl}


\end{document}